\def\year{2022}\relax
\documentclass[letterpaper]{article} 
\usepackage{aaai22}  
\usepackage{times}  
\usepackage{helvet}  
\usepackage{courier}  
\usepackage[hyphens]{url}  
\usepackage{graphicx} 
\usepackage{natbib}  
\usepackage{caption} 
\DeclareCaptionStyle{ruled}{labelfont=normalfont,labelsep=colon,strut=off} 
\usepackage{algorithm}
\usepackage{algorithmic}
\renewcommand{\algorithmiccomment}[1]{\bgroup\hfill \tiny//~#1\egroup}
\usepackage{subfigure}
\usepackage{multirow}
\usepackage{booktabs}
\usepackage{amssymb}
\usepackage{amsmath}

\usepackage{makecell}
\usepackage{comment}
\usepackage{newfloat}
\usepackage{listings}
\floatstyle{ruled}
\newfloat{listing}{tb}{lst}{}
\floatname{listing}{Listing}
\title{Contrastive Continual Learning with Feature Propagation}
\author{
    Xuejun Han,\quad  Yuhong Guo 
}
\affiliations{
Carleton University, Ottawa, Canada\\
    xuejunhan@cmail.carleton.ca, \quad yuhong.guo@carleton.ca
}

\usepackage{bibentry}

\begin{document}

\maketitle

\begin{abstract}
Classical machine learners are designed only to tackle one task without capability of adopting new emerging tasks or classes whereas such capacity is more practical and human-like in the real world. To address this shortcoming, continual machine learners are elaborated to commendably learn a stream of 
	tasks with domain and class shifts among different tasks. In this paper, we propose a general 
	feature-propagation based contrastive continual learning method which is capable of handling multiple 
	continual learning scenarios. 
	Specifically, 
	we align the current and previous representation spaces by means of feature propagation and contrastive representation learning 
	to bridge the domain shifts among distinct tasks. 
	To further mitigate the class-wise shifts of the feature representation,
	a supervised contrastive loss is exploited to make the example embeddings of the same class 
	closer than those of different classes.
	The extensive experimental results demonstrate the outstanding performance of the proposed method on six continual learning benchmarks compared to a group of cutting-edge continual learning methods.
\end{abstract}

\section{Introduction}

In the real world, the environment is not set in stone. The machine learner is desired to favorably response to the changing environment like humans, by acquiring the new knowledge rapidly without forgetting what has learned in the past. Towards this end, continual learning (CL) \cite{CL,LL} came into being to enable the machine learner to learn continuously and has attracted a surge of interest in recent decade because of its pragmatism. Specifically, the continual learner is presented with a stream of non-i.i.d. data and can only learn one task at a time without accessing past task data. Therefore, the major challenge encountered by continual learning is the issue of \emph{catastrophic forgetting} \cite{CF1,CF2} for previously learned tasks. To tackle such problem, a plethora of CL methods were proposed, as well as a variety of evaluation protocols and a systematic categorization of the CL scenarios.

As claimed by \cite{CL2018,Hsu18_EvalCL}, continual learning has three distinct scenarios - task incremental learning, domain incremental learning and class incremental learning - with increasing difficulty. The \emph{task incremental learning} \cite{LWF,SI} is the easiest CL setting where task identifiers are provided at test time and a multi-head network is applied. Explicitly, the unique feature extractor is shared across different tasks but the classifiers are task-specific. By contrast, the \emph{domain incremental learning} \cite{GEM,ER2019} makes use of a single-head network thereby the task identifiers are not required at the test stage. The \emph{class incremental learning} \cite{icarl} as a more realistic but challenging CL scenario learns new tasks continually with a single-head network but the units of the classifier increases with the advent of new classes. 

Distinct from the majority of prior continua learning work which just tackles certain continual learning scenarios, 
in this paper we propose a general feature propagation based contrastive continual learning method
to manage all of them. 
Most existing continual learning methods focus the efforts on retaining either model parameters \cite{EWC,onEWC} or functions \cite{FDR,DERPP} whereas we lay our stress on the representation space. To protect the feature space from drastically changing by aligning the current and past embedding spaces during learning the new task, we desire the model to preserve considerable past knowledge without losing the ability to adapt to new tasks. 
Concretely, 
the method consists of three components. First, 
the current example embeddings are re-represented by being integrated with the previous corresponding ones via feature propagation, 
while experience replay \cite{ER1,ER2} is implemented on such propagated example embeddings. 
Additionally, a contrastive loss is deployed to explicitly enforce the current embeddings to approach the previous ones. 
To further eliminate the domain and classes shifts among distinct tasks, we also use a supervised contrastive loss to 
discriminatively
make the examples of the same class 
closer than those from different classes in the learned embedding space.
Extensive experiments are conducted 
on six standard continual learning benchmarks: 
Split MNIST, Permuted MNIST, Rotated MNIST, Split CIFAR-10, Split CIFAR-100 and Split Tiny ImageNet. 
The results 
demonstrate the superiority of our proposed method 
over a number of 
the state-of-the-art 
continual learning competitors.

\section{Related Work}
\paragraph{Continual learning}
Generally, the continual learning methods can be grouped into three categories. First, \emph{regularization-based methods} 
alleviate  
the issue of forgetting by either regularizing the parameter changes in terms of the importance of parameters for old tasks \cite{EWC,SI,onEWC,RWalk,MAS}, or aligning the current and previous function space by means of knowledge distillation \cite{LWF,icarl} or KL divergence \cite{FDR,DERPP}. Second, \emph{rehearsal-based methods} reserve a memory set of past examples which can be directly enrolled in training as training data \cite{MER,GSS,DERPP,HAL} known as experience replay \cite{ER1,ER2} or indirectly used as constraints \cite{GEM,AGEM} or regularizations \cite{icarl}. It is worth noting that the rehearsal-based methods are 
empirically the state-of-the-art CL methods. Instead of directly storing past examples, \cite{DGR,RtF} train a generative model on previous tasks and rehearsal pseudo-examples during learning new tasks, which is however hard to perform well on complex datasets. Lastly, \emph{model-based methods} \cite{PNN,DEN,MNTP} modify the network dynamically by adding extra units for new tasks or merging a couple of units for similar tasks.
Our proposed method exploits rehearsal memories, while regularizing feature representations in a contrastive manner. 

\paragraph{Contrastive learning}
Contrastive learning has received increasing attention owing to its excellent capacity of learning embedding space, resulting in the 'similar' examples close together and 'dissimilar' examples far apart in the representation space \cite{ctt}. In the case of no label available, the similar examples are generated by data augmentation \cite{SimCLR} or from a feature memory bank \cite{CL2018}. On the contrary, similar examples are the data of the same class and dissimilar examples are data from different classes \cite{svcl}.

\section{Continual Learning Setup}
A continual learner experiences a stream of data triplets $(\mathbf{x}_i, y_i, t_i)$ over time where $t_i\in\{1,\cdots,T \}$ is the task identifier. For each task $t$, 
the i.i.d. examples $(\mathbf{x},y,t)$ are drawn 
from a distribution $\mathcal{D}_t$ whereas the whole data stream is not independently and identically distributed (non-i.i.d.); 
i.e. there are domain and class shifts among different tasks. The continual learner is trained on one task at a time and not able to revisit the data of learned tasks aside from a few pieces of data. The goal of the continual learner is to generally perform well on all learned tasks, namely, rapidly adapting to new tasks and meanwhile preserving the previously learned knowledge to a great extent. Concretely, assuming the learner $g$ has learned up till task $t_c$, the general objective is trying to minimize the following multi-task error:
\begin{equation}
\frac{1}{t_c} \sum_{t=1}^{t_c} \mathbb{E}_{(\mathbf{x},y)\sim\mathcal{D}_t} l(g(\mathbf{x}),y)
\end{equation}
where $l$ is a loss function. The particular challenge faced by continual learning is the problem of \emph{catastrophic forgetting} that means learning new tasks may hurt the performance on past tasks due to the non-i.i.d stream of data. In addition to this challenge, the continual learner is expected to fast acquire and adapt to the new tasks, hence a more compelling setting named \emph{single epoch training} is considered in some prior continual learning work \cite{GEM,AGEM,ER2019} and will be adopted in this paper, where the continual learner can only experience the data stream once. For rehearsal-based CL methods, a small memory set $\mathcal{M}$ storing a few past examples is reserved and can be revisited multiple times along the learning of new tasks. At the time of task t, by incorporating the memory set $\mathcal{M}$ into the current task data $\mathcal{D}_t$ as the training set which is also known as \emph{experience replay} \cite{ER1,ER2}, the objective of the continual learner $g$ is to minimize the following loss:
\begin{equation}
\mathcal{L}_{er} = \mathbb{E}_{(\mathbf{x},y)\sim \mathcal{D}_t \cup \mathcal{M}} l(g(\mathbf{x}),y)
\end{equation}
where $l$ is generally the cross-entropy loss function. It is worth noting that the experience replay is a very simple and strong CL baseline \cite{ER2019} and outperforms significantly in terms of either performance or efficiency the rehearsal-based methods that utilize the memory set indirectly during training such as \cite{icarl,GEM}. Therefore, our proposed method is developed on the basis of experience replay.

\begin{figure}[t]
\centering
\includegraphics[width=0.99\columnwidth]{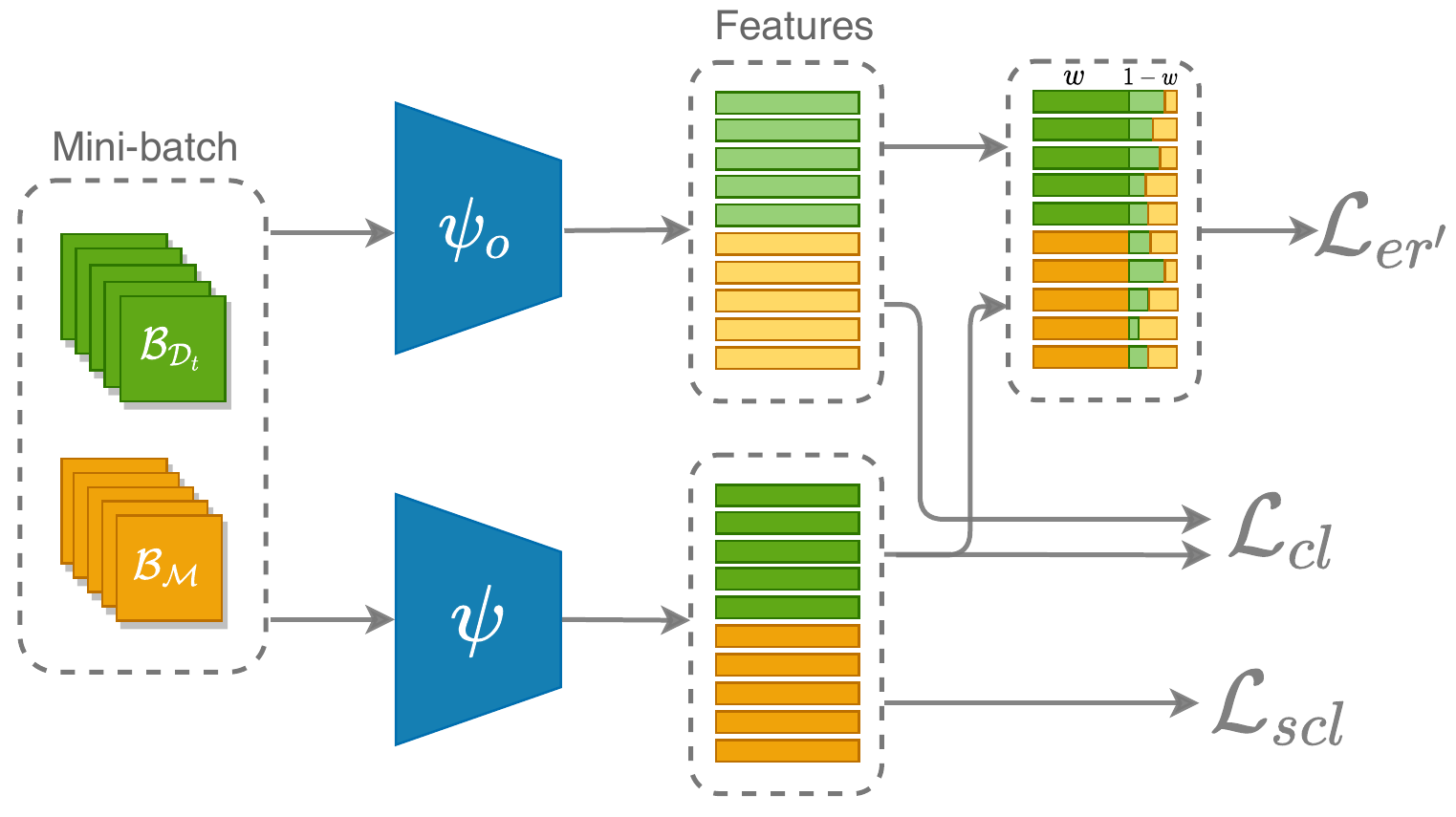} 
\caption{An overview of the proposed method CCL-FP+. $\psi_{o}$ is the feature extractor whose state is after learning task $t-1$ but before task $t$, and remains fixed during the training. $\psi$ is the feature extractor that is being optimized. The minibatch for each training iteration in generted by concatenating a minibatch $\mathcal{B}_{\mathcal{D}_t}$ from the current task data $\mathcal{D}_t$ with another minibatch $\mathcal{B}_\mathcal{M}$ from the memory set $\mathcal{M}$.}
\label{fig:framework}
\end{figure}

\section{Methodology}
Instead of enforcing the model outputs to be close to the previous ones in the function space by means of knowledge distillation loss in some prior continual learning work \cite{LWF,icarl,DERPP}, we turn to  focus on the representation space to preserve it from drastic changes by means of feature propagation and contrastive loss. Specifically, the current representation space first absorbs some past knowledge by feature propagation from previous feature space. A contrastive regularization
loss is then employed to reinforce such effect by encouraging the current example embeddings to approach the previous ones. As a complement, a supervised contrastive loss is leveraged to push the examples of the same class to cluster tightly in the representation space whereas the examples' embeddings from different classes are driven to be far apart. 
An overview of the proposed method is illustrated in Figure \ref{fig:framework}.
We will give an exposition in following sections

\subsection{Experience Replay with Feature Propagation}
Suppose the model consists of two components: the feature extractor $\psi$ and the classifier $f$. It is noted that for class incremental learning and domain incremental learning, there is an only classifier for all tasks whereas for task incremental learning, there is one classifier for each task. The feature extractor 
remains unique all the time. When the task $t$ arrives, we have two labelled datasets on hand: the current task data $\mathcal{D}_t$ and a small memory set $\mathcal{M}$ of a few past examples. Before retraining the model on new data, we first copy the feature extractor $\psi$ to $\psi_o$ and keep $\psi_o$ fixed during the ensuing training. Afterwards, a feature propagation procedure is carried out for the training data $\mathcal{D}_t\cup\mathcal{M}$ in the representation space. Concretely, the example embeddings derived from $\psi$ are amended by fusing a weighted sum of all 
example embeddings come from $\psi_o$. We denote the modified example embeddings by $\tilde{\psi}(\mathbf{x})$ defined as follows,
\begin{equation}
\tilde{\psi}(\mathbf{x}_i)=(1-w)\cdot \psi(\mathbf{x}_i) + w\cdot \sum_j A_{ij}\psi_{o}(\mathbf{x}_j) 
\end{equation}
where $w\in(0,1)$ is a trade-off parameter to balance the current and previous embedding spaces. The propagation weight $A_{ij}$ for examples $\mathbf{x}_i$ and $\mathbf{x}_j$ is set as:
\begin{equation}
A_{ij} = \frac{\exp(-d(\psi(\mathbf{x}_i), \psi_{o}(\mathbf{x}_j)) \cdot \eta)}  {\sum_{\mathbf{x}_{j'}\in\mathcal{D}_t\cup\mathcal{M}} \exp(-d(\psi(\mathbf{x}_i), \psi_{o}(\mathbf{x}_{j'})) \cdot \eta )} 
\end{equation}
where $d(\cdot,\cdot)$ is the Euclidean distance and $\eta$ is a temperature parameter. In consequence, the amended experience replay loss is 
\begin{equation}
\mathcal{L}_{er'} = \mathbb{E}_{(\mathbf{x},y)\sim \mathcal{D}_t \cup \mathcal{M}} l(f(\tilde{\psi}(\mathbf{x})),y)
\end{equation}
where $l$ is the cross-entropy loss function. By being trained on the basis of the fused representation space, the model is expected to absorb certain propagated global 
information from the previous embedding space so as to avoid the significant forgetting issue.

\subsection{Contrastive Representation Rehearsal}
The contrastive learning shows the excellent capacity for representation learning by pushing the 'similar' examples to be close together and 'dissimilar' examples to be far apart \cite{CL2018,SimCLR,MoCo}. Inspired by such idea but distinct from some existing work such as \cite{CL2018} where the 'similar' examples are formed from the representation space of the previous training iteration, our model makes use of the example embeddings derived from the previous feature extractor $\psi_o$. Explicitly, we enforce the example embeddings to stay near the previous 
corresponding ones by a contrastive loss, so that to realize the aspiration of protecting the representation space from dramatically changing after being retrained on new tasks. The proposed contrastive loss is defined as follows,
\begin{equation}
\begin{split}
\mathcal{L}_{cl} = &- \mathbb{E}_{\mathbf{x}\sim \mathcal{D}_t \cup \mathcal{M}} \\
&  \log \frac{\exp(-d({\psi}(\mathbf{x}), \psi_{o}(\mathbf{x}))\cdot \tau)}  {\sum_{\mathbf{x}_{j}\in\mathcal{D}_t\cup\mathcal{M}}  \exp(-d({\psi}(\mathbf{x}), \psi_{o}(\mathbf{x}_{j})) \cdot \tau )}
\end{split}
\end{equation}
where $d(\cdot,\cdot)$ is the Euclidean distance and $\tau$ is a temperature parameter. It is noted that instead of merely applying the contrastive learning to memory data $\mathcal{M}$ which the model has previously seen and learned, the whole training data $\mathcal{D}_t\cup\mathcal{M}$ is deployed in this loss to enhance the contrastive effect which we empirically found performs better.

Up to this point, we derive our proposed method which we named as \emph{contrastive continual learning 
with feature propagation (CCL-FP)}. The overall objective during task $t$ is 
\begin{equation}
\mathcal{L}_{ccl-fp} = \mathcal{L}_{er'} + \alpha\mathcal{L}_{cl}
\end{equation}
where $\alpha\in(0,1)$ is a trade-off parameter. By minimizing the above objective, the representation space is expected to stay steady
throughout the training of all tasks, which is intuitively adequate to cope with the task incremental learning where the model is equipped with task specific classifiers or the class incremental learning in the case that the classes for different tasks are disjoint. Nevertheless, for the domain incremental learning where all tasks share an only classifier, it may be reliable to a limited degree.

\begin{table*}[t]
    \centering
     \setlength\tabcolsep{2.5pt}
    \begin{tabular}{lccccc} 
        \toprule
        \multirow{1}{*}[-1em]{\textbf{Model} } & \multicolumn{2}{c}{ \textbf{S-MNIST}} &  \multicolumn{2}{c}{\textbf{S-CIFAR-10}}   & \textbf{P-MNIST}  \\
          \cmidrule(lr){2-3}  \cmidrule(lr){4-5} \cmidrule(lr){6-6}
         {} & Class-IL & Task-IL &Class-IL & Task-IL & Domain-IL\\
        \midrule
        Joint &$ \mathbf{95.59\pm0.31} $    &$ \mathbf{99.33\pm0.17} $      &$ \mathbf{58.89\pm3.26} $     &$\mathbf{87.58\pm1.85} $ 	  &$ \mathbf{77.65\pm1.09 } $     \\
        Finetune &$19.62 \pm0.12 $    &$ 95.25\pm1.66 $      &$ 17.00\pm 1.20$     &$64.02 \pm 3.53$ 	  &$ 58.68\pm 0.46$     \\
         \cmidrule(lr){1-6}     
	  SI \cite{SI} &$ 26.16\pm0.75 $   &$ 96.56\pm 0.58 $    &$17.14 \pm0.13 $    &$ 63.31\pm 3.79$ 	  &$ 58.31\pm2.76 $     \\  
	  LwF \cite{LWF} &$21.30\pm 0.92$     &$ 98.84\pm 0.19$     &$ 20.82\pm 0.35 $   &$ 60.41\pm 2.89$ & --  \\      
	  ER-Res. \cite{ER2019} &$76.43\pm 3.08$     &$ 98.77\pm 0.14$     &$ 44.45\pm3.69 $   &$ 84.42\pm 1.15$ & $66.95\pm1.40$  \\    
       GEM \cite{GEM}  &$80.79\pm 1.47$     &$ 97.68\pm 0.32$     &$ 18.66\pm 0.91$     &$ 77.74\pm 2.60$ 	  &$ 62.96\pm1.14 $        \\
	  A-GEM \cite{AGEM} &$ 45.69\pm 3.77$   &$98.66 \pm 0.16$    &$ 18.13\pm 0.27$    &$ 74.07\pm0.76 $ 	  &$ 60.48\pm 2.04$    \\
  iCaRL \cite{icarl} &$ 69.95\pm 0.07$   &$98.30 \pm 0.11$    &$ 36.68\pm 1.80$    &$ 82.24\pm 0.64 $ 	  &--    \\   
FDR \cite{FDR} &$ 81.03\pm 2.23$    &$ 98.66\pm0.52 $    &$ 19.51\pm 1.04$     &$ 74.29\pm 3.49 $      &$ 68.41\pm 2.72 $   \\
	  HAL \cite{HAL} &$ 79.15\pm2.03 $     &$ 98.81\pm 0.18$       &$ 33.86\pm 1.73$     &$ 75.19\pm 2.57$     &$\mathbf{70.83 \pm 1.86}$    \\
	   \cmidrule(lr){1-6}   
	  CCL-FP (ours) &$ 88.67\pm 0.97$   &$ \mathbf{99.15\pm 0.37}$   &$ 50.11\pm3.69 $    &$ 85.44\pm 2.03$ 	  &$66.91 \pm0.95 $      \\
	  CCL-FP+ (ours) &$\mathbf{89.16} \pm 1.14$  &$ 99.14\pm 0.05$     &$\mathbf{51.74 \pm 2.41}$     &$\mathbf{86.33 \pm 1.47 }$ 	  &$ 69.22\pm1.07 $      \\
        \bottomrule
    \end{tabular}
        \bigskip
       
    \begin{tabular}{lccccc} 
        \toprule
        \multirow{1}{*}[-1em]{\textbf{Model} } & \multicolumn{2}{c}{ \textbf{S-CIFAR-100}} &  \multicolumn{2}{c}{\textbf{S-Tiny-ImageNet}}   & \textbf{R-MNIST}  \\
          \cmidrule(lr){2-3}  \cmidrule(lr){4-5} \cmidrule(lr){6-6}
         {} & Class-IL & Task-IL &Class-IL & Task-IL & Domain-IL\\
        \midrule
        Joint &$\mathbf{19.60\pm2.14} $    &$ \mathbf{69.80\pm2.17} $      &$ \mathbf{14.21\pm0.66} $     &$\mathbf{43.89 \pm 0.88}$ 	  &$ \mathbf{84.12\pm0.61} $     \\
        Finetune  &$ 3.58\pm0.13 $   &$ 39.55\pm4.42 $    &$4.77 \pm0.23 $    &$ 26.93\pm 1.59$ 	  &$ 67.64\pm 2.17$   \\
         \cmidrule(lr){1-6}     
	  SI \cite{SI} &$ 3.55\pm 0.08$   &$ 36.33\pm 4.23$    &$ 3.77\pm 0.52$    &$ 17.85\pm 0.92$ 	  &$71.44\pm0.33  $     \\
	    Lwf. \cite{LWF}  &$ 3.86\pm 0.18$     &$ 30.06\pm 1.52$  &$ 5.64\pm 0.12$      &$14.68 \pm 1.20$   & --   \\          
	  ER-Res. \cite{ER2019} &$ 9.74\pm 0.98$   &$ 63.05\pm 0.82$    &$ 7.21\pm 0.29$     &$36.75 \pm 0.79$   &$ 79.77\pm 0.86 $    \\   
	         GEM \cite{GEM}  &$ 4.69\pm 0.41 $     &$ 49.29\pm 0.73$     & $6.76\pm0.45$      &$29.05\pm0.74$  	  &$79.85 \pm 2.17$        \\
	  A-GEM \cite{AGEM} &$ 3.67\pm 0.10$   &$ 46.88\pm1.81 $    &$ 5.43\pm0.11 $    &$29.67 \pm 0.91$ 	  &$ 73.64\pm 3.68$    \\
 iCaRL \cite{icarl}   &$ 9.64\pm 0.25$    &$ 53.49\pm 0.88 $     & $3.84\pm 0.27$      &$19.49\pm1.17$  	  &--       \\    
FDR \cite{FDR} &$ 3.65\pm 0.10$    &$ 42.87\pm 2.62$    & $4.83\pm0.43$      &$26.97\pm 2.69$     &$79.75\pm 3.12$   \\
	  HAL \cite{HAL} &$ 6.31\pm0.71 $     &$47.88 \pm 2.76$       & $3.85\pm0.32$      &$21.70\pm1.12$  	  &$ 78.65\pm1.57 $    \\
	   \cmidrule(lr){1-6}   
	  CCL-FP (ours) &$13.64 \pm1.04 $   &$\mathbf{65.19 \pm0.65 }$   &$ \mathbf{10.52\pm0.28} $    &$ 39.44\pm 0.48$ 	  &$ 80.68\pm 1.74$      \\
	  CCL-FP+ (ours) &$ \mathbf{14.05\pm 0.85}$  &$ 65.19\pm1.88  $     &$ 10.13\pm0.44 $     &$ \mathbf{39.99\pm 0.59} $ 	  &$ \mathbf{82.06\pm1.29}$      \\
        \bottomrule
    \end{tabular}
\caption{The average accuracy $\pm$ standard deviation (\%) by the end of training for baselines and our models across 5 runs with different random seeds. The results for joint training, i.e. the upper bound, and the best accuracies for CL models on each benchmark are marked in bold. It is noted that '--' indicates experiments are unable to run because of compatibility issues (e.g. LwF and iCaRL in domain incremental learning). }
 \label{tab:acc}
\end{table*}

\subsection{Supervised Contrastive Replay}
In the domain incremental learning, the classes for each task are identical whereas the domain shifts among different tasks are relatively substantial. In this case, merely retaining the representation space may not be 
a panacea, especially as
all tasks share a unique classifier. Out of such concern, we resort to supervised contrastive learning as a complement to our overall loss function. Since the labels are on hand in our continual learning setup, the `similar' examples here can be formed by the data of the same class from $\mathcal{D}_t\cup\mathcal{M}$. Hence, a supervised contrastive loss is defined as follows, 
\begin{equation}
\begin{split}
\mathcal{L}_{scl}= &- \mathbb{E}_{\mathbf{x}_i\sim \mathcal{D}_t \cup \mathcal{M}}   \mathbb{E}_{\mathbf{x}_k\sim\mathcal{S}_i }  \\
&\log \frac{\exp(-d({\psi}(\mathbf{x}_i), \psi_{}(\mathbf{x}_k)) \cdot \tau)}  {\sum_{\mathbf{x}_j\in(\mathcal{D}_t\cup\mathcal{M}) \backslash\mathbf{x}_i} \exp(-d({\psi}(\mathbf{x}_i), \psi_{}(\mathbf{x}_{j})) \cdot \tau )}
\end{split}
\end{equation}
where $d(\cdot,\cdot)$ is the Euclidean distance and $\tau$ is a temperature parameter. $\mathcal{S}_i$ is a set consisting of the examples from $\mathcal{D}_t \cup \mathcal{M}$ of the same class as $\mathbf{x}_i$ but excluding $\mathbf{x}_i$ itself. The model thereby learns to make example embeddings of the same class which may come from both $\mathcal{D}_t$ and $\mathcal{M}$ close together and those of different classes distinct, so that the domain shifts of the same class among different tasks can be eliminated if there is any. 

By integrating the supervised contrastive loss into the overall objective, we obtain our intensified model named by \emph{CCL-FP+}:
\begin{equation}
\mathcal{L}_{ccl-fp+} =  \mathcal{L}_{er'} + \alpha\mathcal{L}_{cl} + \beta\mathcal{L}_{scl}
\end{equation}
where  $\alpha$ and $\beta$ are trade-off parameters in the range of $(0,1)$. We solve it by using a batch-wise gradient descent algorithm. 
The training algorithm for CCL-FP+ is provided in the supplementary file, 
while the memory set $\mathcal{M}$ is updated along the training by reservoir sampling \cite{Reservoir}.

\begin{figure*}[t]
    \centering
    \subfigure[Split CIFAR-10 $|$ Class-IL]{
        \includegraphics[width=0.5\textwidth]{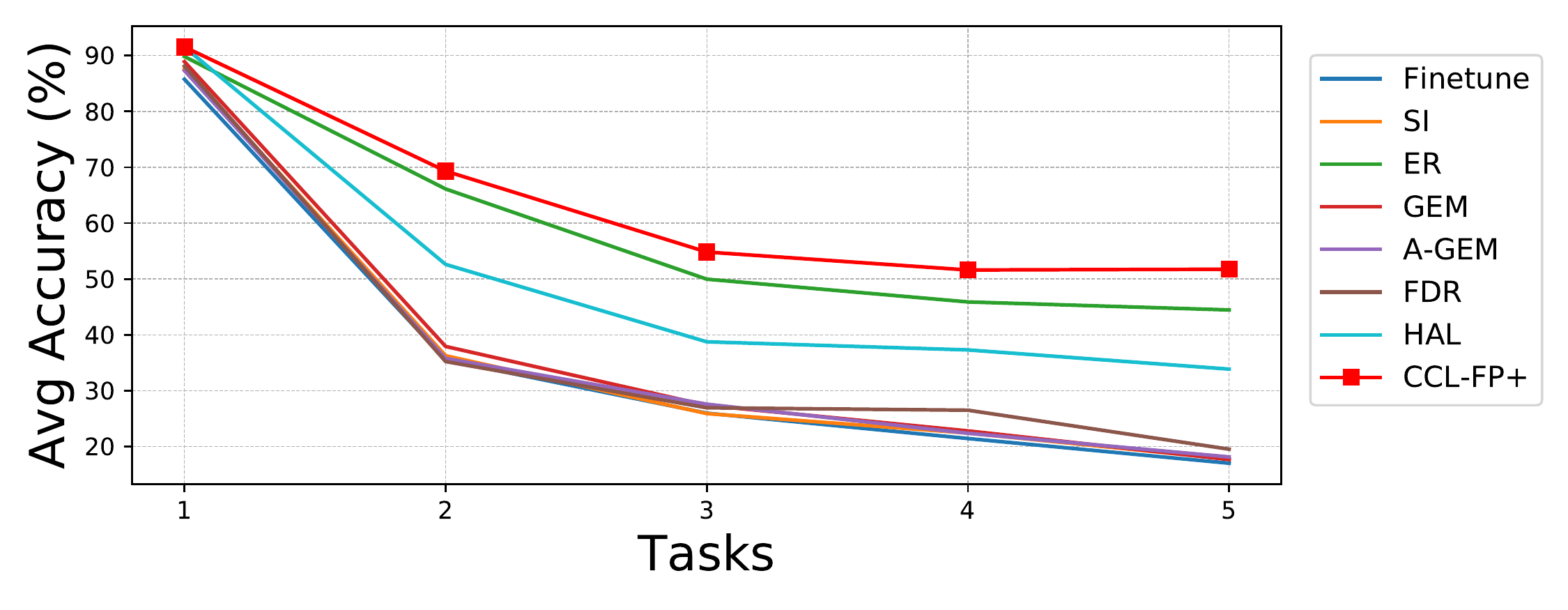}
    }
     \hspace{-0.75in}
    \subfigure[Split CIFAR-10 $|$ Task-IL]{
	\includegraphics[width=0.5\textwidth]{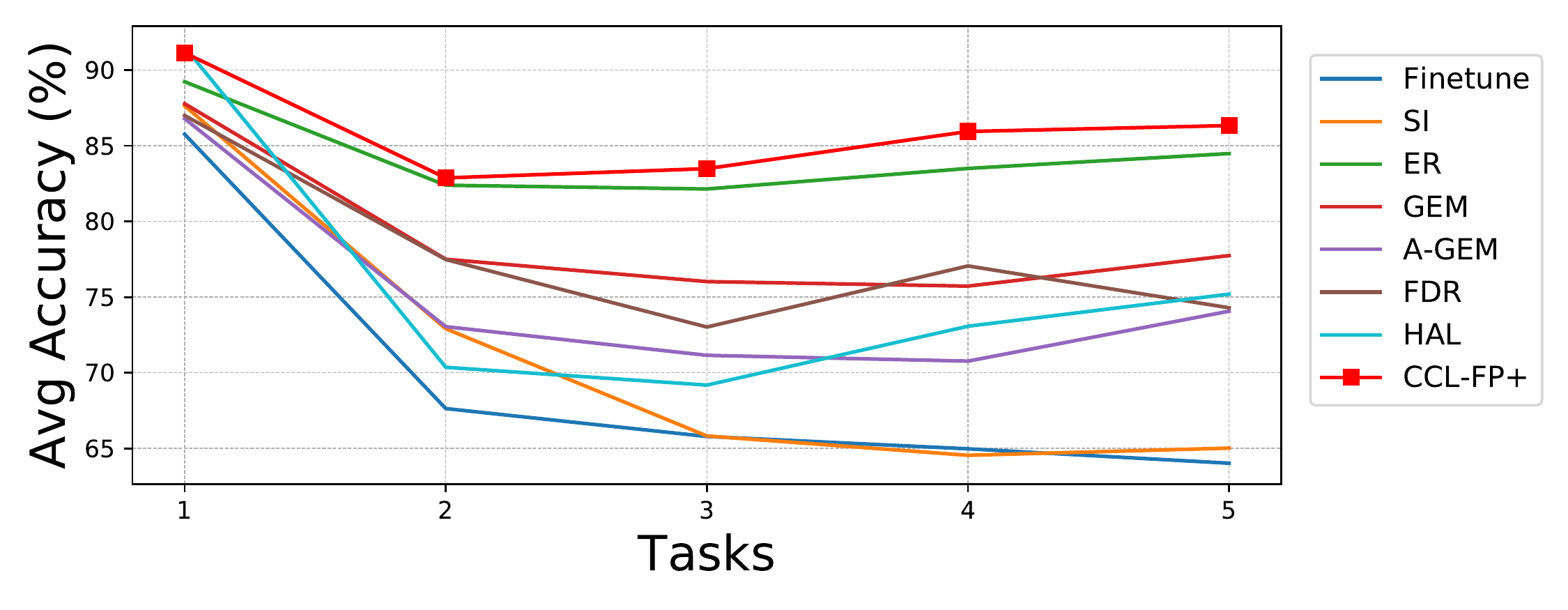}
    }
    \quad    
    \subfigure[Split CIFAR-100 $|$ Class-IL]{
    	\includegraphics[width=0.5\textwidth]{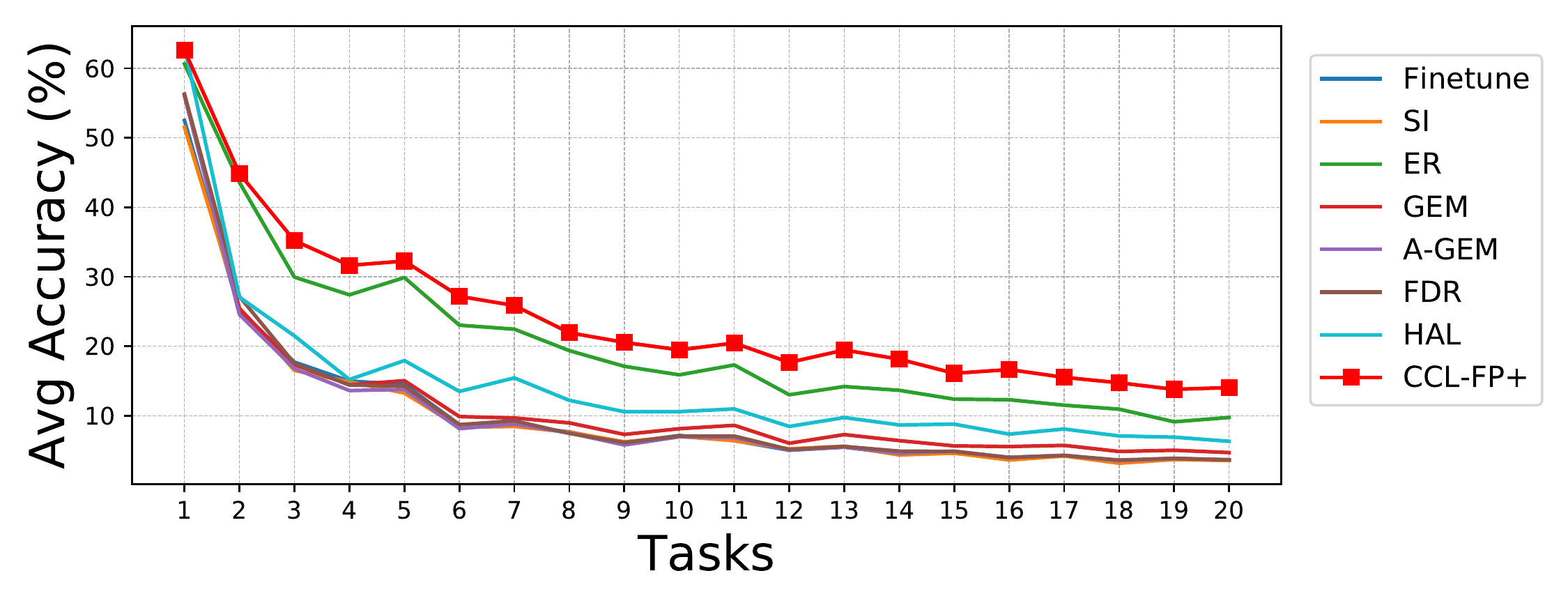}
    }
     \hspace{-0.75in}
    \subfigure[Split CIFAR-100 $|$ Task-IL]{
	\includegraphics[width=0.5\textwidth]{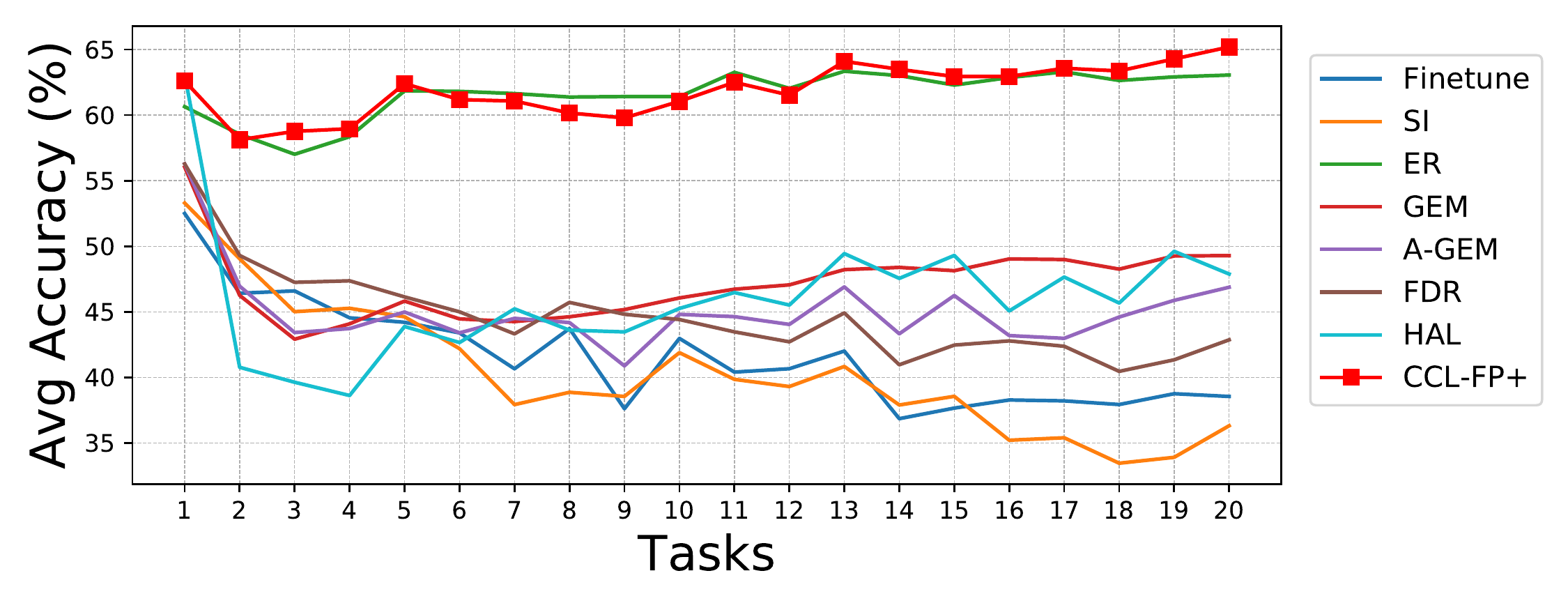}
    }
    \quad    
    \subfigure[Permuted MNIST  $|$ Domain-IL]{
    	\includegraphics[width=0.5\textwidth]{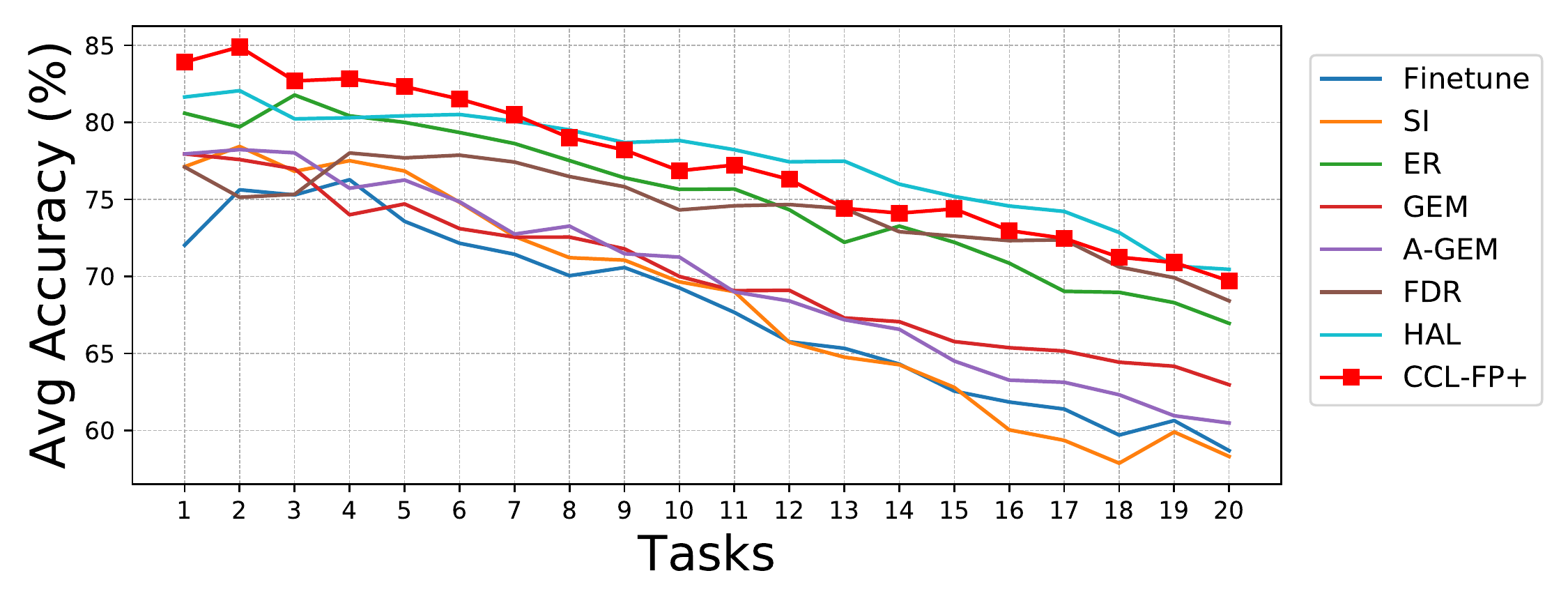}
    }
     \hspace{-0.75in}
    \subfigure[Rotated MNIST  $|$ Domain-IL]{
	\includegraphics[width=0.5\textwidth]{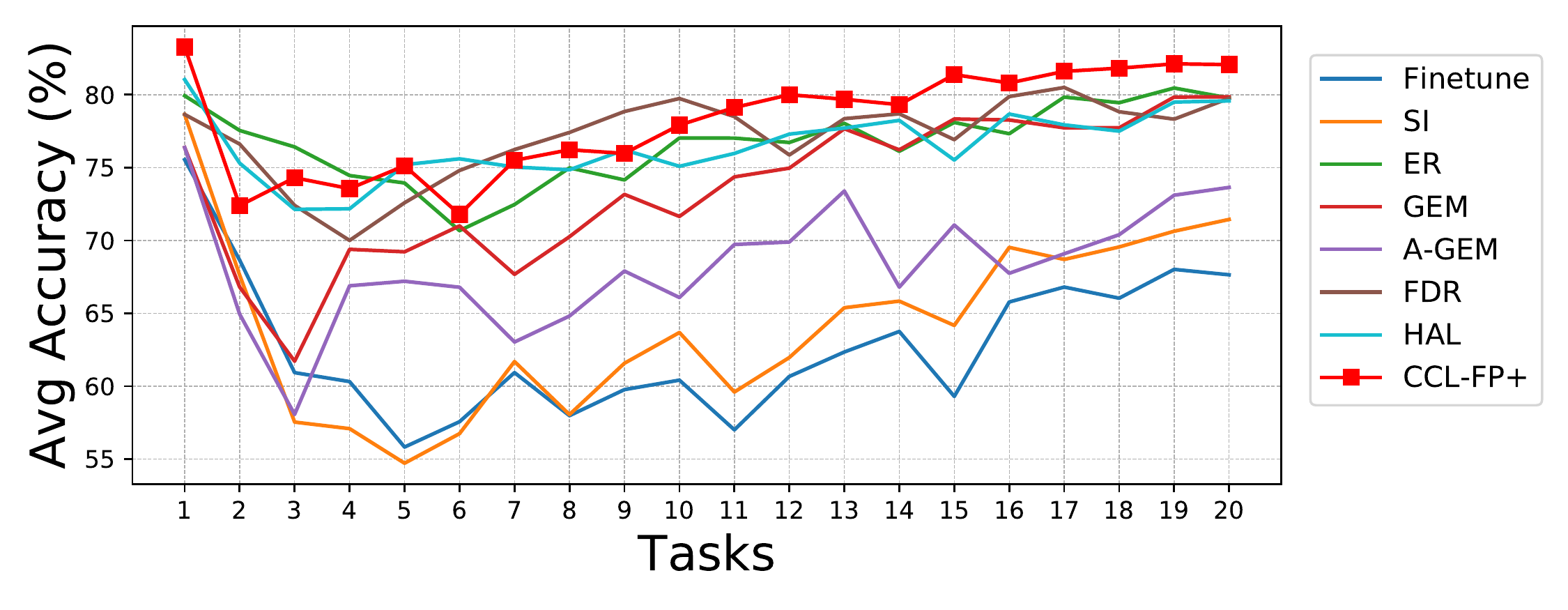}
    }
    \caption{The evolution of average accuracy on test data of all seen tasks as new tasks are learned on selected datasets and settings. All results are obtained across 5 runs with different random seeds.}
    \label{fig:evolution}
\end{figure*}

\begin{figure}[t]
\centering
\includegraphics[width=0.99\columnwidth]{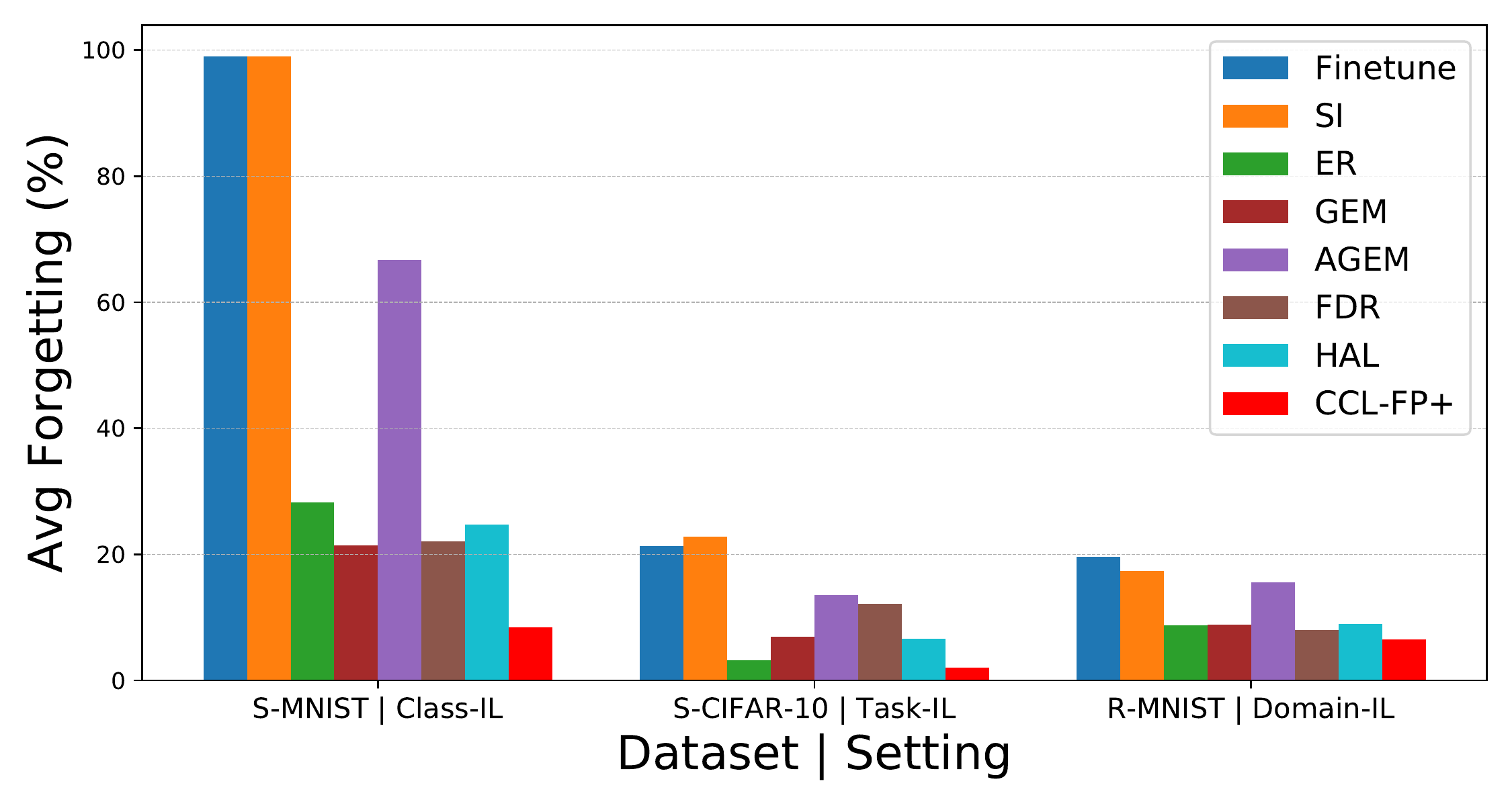} %
\caption{The average forgetting (\%) for baselines and our model across 5 runs with different random seeds on Split MNIST (class-il), Split CIFAR-10 (task-il) and Rotated MNIST (domain-il). 
	}
\label{fig:forgetting}
\end{figure}

\begin{figure}[t]
\centering
\includegraphics[width=0.98\columnwidth]{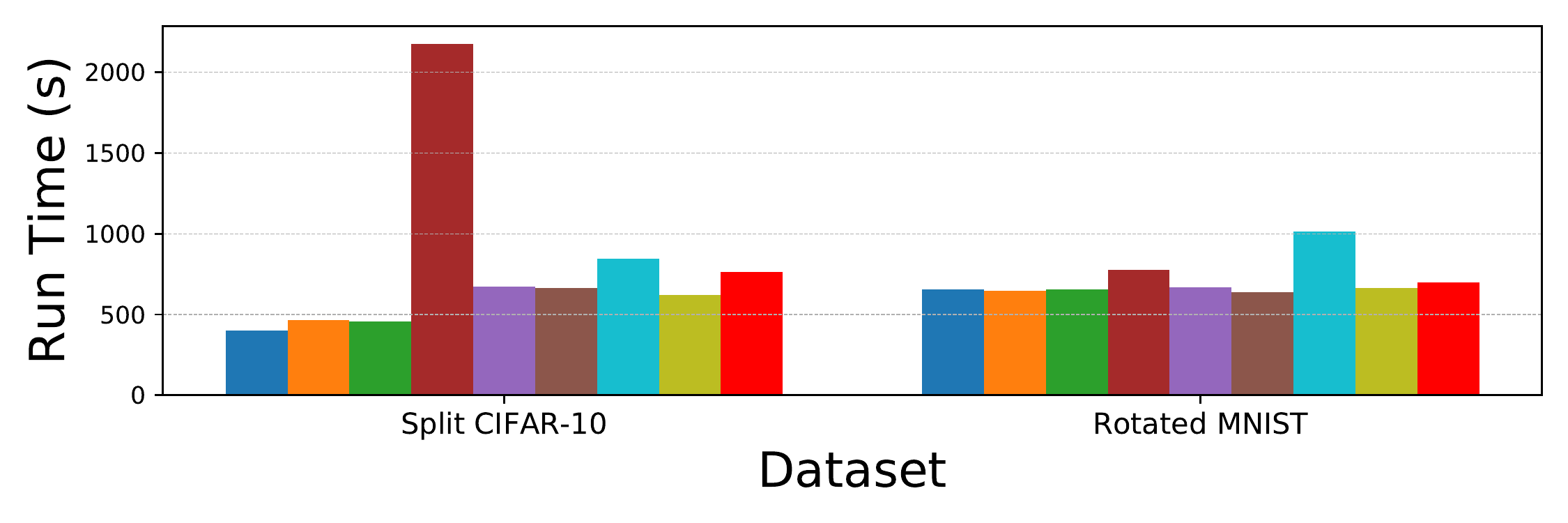} 
\includegraphics[width=0.89\columnwidth]{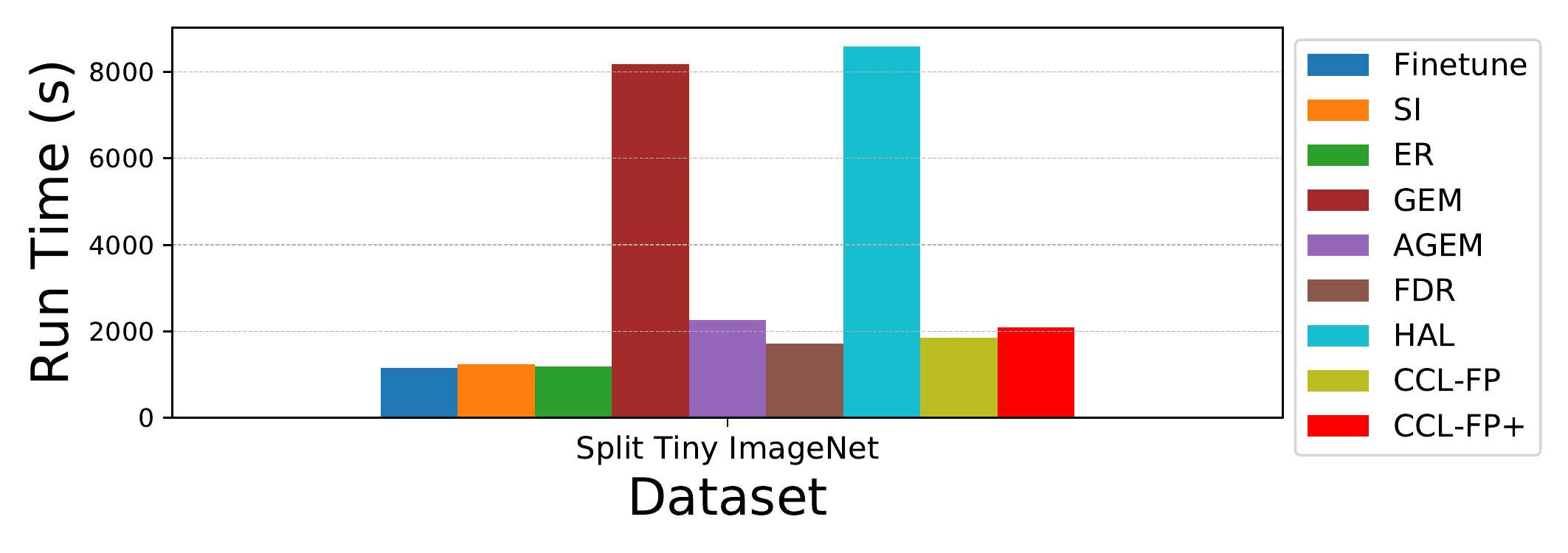} 
\caption{Run time (training + inference) for all baselines and our model on three selected datasets - Split Cifar-10, Splite Tiny-ImageNet and Rotated MNIST.}
\label{fig:runtime}
\end{figure}

\begin{table*}[tb]
    \centering
     \setlength\tabcolsep{8pt}
    \begin{tabular}{llccccc} 
        \toprule
        \multirow{1}{*}[-1em]{\textbf{Buffer} } & \multirow{1}{*}[-1em]{\textbf{Model} } & \multicolumn{2}{c}{ \textbf{S-CIFAR-10}} &  \multicolumn{2}{c}{\textbf{S-CIFAR-100}}  & \textbf{R-MNIST}  \\
          \cmidrule(lr){3-4}  \cmidrule(lr){5-6} \cmidrule(lr){7-7} 
        {}& {} & Class-IL & Task-IL &Class-IL & Task-IL & Domain-IL  \\
        \midrule
        200 &ER-Res.	&$44.45\pm3.69$  	&$84.42\pm1.15$	&$9.74\pm0.98$	&$63.05\pm0.82$	&$79.77\pm0.86$\\
        		&HAL  &$33.86\pm1.73$  	&$75.19\pm2.57$	&$6.31\pm0.71$	&$47.88\pm2.76$	&$78.65\pm1.57$\\
		 \smallskip 
		&CCL-FP+ & $51.74\pm2.41$    &$ 86.33 \pm1.47$      &$ 14.05 \pm0.85$     &$65.19 \pm 1.88$ 	 &$ 82.06 \pm 1.29$ \\
        500  &ER-Res.	& $56.64\pm1.56$ 	&$87.02\pm0.63$	&$14.55\pm1.51$	&$68.87\pm0.73$	&$82.23\pm0.91$\\
	  	&HAL&$37.48\pm5.07$  	&$75.87\pm3.77$	&$8.09\pm1.29$	&$50.04\pm4.60$	&$82.30\pm0.35$\\
		 \smallskip 
		&CCL-FP+  &$ 57.03\pm3.42 $   &$ 87.37\pm1.09 $    &$19.44\pm0.54  $    &$ 69.03\pm0.67$ 	   &$ 83.10\pm1.39$\\   
	  1000  &ER-Res.	&$57.19\pm2.99$  	&$88.35\pm2.38$	&$21.26\pm0.69$	&$73.30\pm0.78$	&$84.48\pm1.80$\\
	  	&HAL 	&$45.13\pm3.63$  	&$80.87\pm2.42$	&$11.02\pm1.25$	&$56.59\pm3.21$	&$84.55\pm0.96$\\
		 \smallskip 
		&CCL-FP+  &$ 60.29\pm4.15$   &$ 88.87\pm1.88$    &$ 23.99\pm1.93$    &$ 74.27\pm0.85$ 	  &$ 84.95\pm1.03$  \\
	  5000&ER-Res. &$61.71\pm5.49$  	&$90.94\pm0.71$	&$24.27\pm1.18$	&$79.71\pm0.60$	&$85.87\pm1.15$\\
	  	&HAL & $48.83\pm4.41$ 	&$83.35\pm2.63$	&$15.31\pm2.49$	&$65.40\pm3.85$	&$85.46\pm 1.56$\\
		&CCL-FP+	&$ 63.84\pm4.95$   &$ 91.03\pm1.48$    &$ 27.38\pm1.32$     &$80.46\pm1.88 $    &$ 86.96\pm0.91$\\       
        \bottomrule
    \end{tabular}
  \caption{The average accuracy $\pm$ standard deviation (\%) across 5 runs with different random seeds over a range of  buffer size of ER, HAL and CCL-FP+ on three selected benchmarks - Split CIFAR-10, Split CIFAR-100 and Rotated MNIST.}
  \label{tab:buffer}
\end{table*}

\section{Experiments}
In this section, we compare 
our models CCL-FP and CCL-FP+ with eight continual learning methods on six benchmark datasets over three continual learning settings -- class incremental learning, task incremental learning and domain incremental learning.

\subsection{Experimental Setting}
\paragraph{Datasets}
We conducted extensive experiments on six commonly used benchmarks in the continual learning literature, of which four are applied in class and task incremental learning and the other two are in domain incremental learning. \textbf{Split MNIST} \cite{SI} is constructed by splitting the source MNIST dataset \cite{mnist} into 5 disjoint binary-class subsets in sequence (e.g. 0/1, 2/3, 4/5, 6/7, 8/9), of which each is considered as a separate task. \textbf{Permuted MNIST} \cite{EWC} is a variant of the 
MNIST dataset, where each task applies a certain random pixel-level permutation to all the original images. \textbf{Rotated MNIST} \cite{GEM} is another variant of MNIST by rotating the original images with 
a certain random angel between 0 and 180 degrees in each task. For Permuted MNIST and Rotated MNIST, we consider 20 tasks and each task has 1000 examples of 10 classes randomly sampled from the entire dataset. \textbf{Split CIFAR-10} \cite{SI} and \textbf{Split CIFAR-100} \cite{icarl} are constructed by splitting the CIFAR-10 and CIFAR-100 datasets \cite{cifar10} into 5 disjoint binary-class subsets and 20 disjoint 5-class subsets, respectively. Similarly, \textbf{Split Tiny ImageNet} is a sequential split of the original Tiny ImageNet dataset \cite{tinyimg} with 10 tasks, each of which introduces 20 classes. 

\paragraph{Baselines}
We compared the proposed methods CCL-FP and CCL-FP+ with eight CL competitors as well as the upper bound and lower bound. \textbf{Joint} trains the model with access to the data of all tasks at the same 
time, 
which serves as an upper bound in terms of performance. \textbf{Finetune} is a lower bound for CL baselines which simply trains the model using new task data without any effort to overcome forgetting. \textbf{SI} \cite{SI} is a regularization-based CL method by applying a penalty to model parameters. \textbf{LwF} \cite{LWF} is another regularization-based model with a regularizer in the functional space by leveraging knowledge distillation.
The rest of CL baselines are categorized into rehearsal-based methods. \textbf{ER-Reservoir} \cite{ER2019} conducts 
experience replay with a memory set updated by reservoir sampling \cite{Reservoir}. \textbf{GEM} \cite{GEM} exploits the memory set as multiple constraints to project the parameter gradients so as to overcome forgetting. \textbf{A-GEM} \cite{AGEM} is a simplified and efficient version of GEM. \textbf{iCaRL} \cite{icarl} makes use of a distillation loss on previous tasks to overcome forgetting and performs classification with a nearest-mean-of-exemplars classifier.
\textbf{FDR} \cite{FDR} measures and regularizes function distance in a $L^2$ Hilbert space so as to minimally interferes with what has previously learned. \textbf{HAL} \cite{HAL} complements experience replay with bilevel optimization to regularize the training objective.

\paragraph{Implementation}
All baselines and our models adopt the same backbone architectures. We use a fully-connect network with two hidden layers of 100 RELU units for the variants of the MNIST dataset following \cite{GEM,MER} and a ResNet18 for Split CIFAR-10, Split CIFAR-100 and Split Tiny ImageNet datasets following \cite{icarl,DERPP}. All experiments are implemented under the single epoch setting with minibatch size of 10 \cite{GEM,ER2019}. For temperature parameters $\eta$ and $\tau$, we set $\eta,\tau=0.1$ for datasets in the class and task incremental learning, i.e. Split MNIST, Split CIFAR-10, Split CIFAR-100 and Split Tiny ImageNet, and $\eta=0.1,\tau=1$ for datasets in domain incremental learning, i.e. Permuted MNIST and Rotated MNIST. The hyperparameters $w$ are selected from $\{0.1,0.3,0.5\}$ and $\alpha, \beta$ are from $\{0.01,0.1,0.5,1 \}$. The buffer size $|\mathcal{M}|$ is set to be 200 across all experiments of rehearsal-based CL methods. It is noted that the buffer of size 200 is fairly small and makes the rehearsal-based continual learning difficult especially on Split CIFAR-100 and Split Tiny ImageNet datasets with 200 classes in total.

\begin{table}[t]
    \centering
         \setlength\tabcolsep{2pt}
\resizebox{\columnwidth}{!}{  	 
    \begin{tabular}{cccccccc} 
        \toprule
    {}   & {}  &{} & \multicolumn{2}{c}{ \textbf{S-MNIST}} &  \multicolumn{2}{c}{\textbf{S-CIFAR-10}}   & \textbf{R-MNIST}   \\
          \cmidrule(lr){4-5}  \cmidrule(lr){6-7} \cmidrule(lr){8-8} 
	    \makecell{$w$\\$\neq$\\ $0$} & \makecell{$\alpha$\\$\neq$\\ $0$}  & \makecell{$\beta$\\$\neq$\\ $0$} & Class-IL & Task-IL &Class-IL & Task-IL & Domain-IL  \\
        \midrule
        & & &$76.43$ &$98.77$	&$44.45$	&$84.42$	&$79.77$\\
       \checkmark &   & {}   &$ 83.48 $      &$ 98.72 $     &$50.45 $ 	  &$ 85.23 $    &$ 79.38 $ \\
        &\checkmark    &    &$ 78.04 $    &$99.13  $    &$ 44.82$ 	  &$ 84.63$   &$ 79.89 $\\   
	&    &\checkmark   &$ 77.37$    &$ 98.67$    &$ 46.41$ 	  &$85.55  $   &$  81.01$  \\
	 \checkmark   &\checkmark  & &$ 88.67$   &$ 99.15$    &$ 50.11$     &$85.44 $   &$ 80.68$    \\    
	  \checkmark &    &\checkmark   &$ 82.95$   &$ 98.73$    &$ 50.36$     &$84.73 $   &$ 81.78$    \\       
	    & \checkmark   &\checkmark   &$ 78.31$   &$ 98.89$    &$ 47.52$     &$85.91 $   &$ 81.79$    \\      
	   \checkmark    & \checkmark   &\checkmark   &$ 89.16$   &$ 99.14$    &$ 51.74$     &$86.33 $   &$ 82.06$    \\    
        \bottomrule
    \end{tabular}
	}	
  \caption{The ablation study on three selected benchmarks - Split MNIST, Split CIFAR-10 and Rotated MNIST. All results are the average accuracy across five runs with different random seeds. The results in the first row is for ER with reservoir buffer.}
  \label{tab:ablation}
\end{table}

\paragraph{Metrics}
For a principled evaluation, we consider two metrics, \emph{average accuracy} and \emph{forgetting measure} \cite{RWalk}, to evaluate the performance of models on test data of all tasks. 
The average accuracy evaluates the overall performance for all seen tasks, while 
the forgetting measure reflects the accuracy drops on previous tasks after the model is trained on new tasks. 
The large forgetting values signify the model has less stability when encountering new tasks.

\subsection{Experimental Results}
The overall average accuracy of baselines and proposed models on all benchmarks is reported in Table \ref{tab:acc} and the evolution curves of average accuracy with respect to the number of tasks on selected benchmarks are shown in Figure \ref{fig:evolution}. It is worth noting that our models achieve the best average accuracy on all datasets under different settings, except Permuted MNIST in which HAL behaves better, whereas our models are more computational efficient as shown in Figure \ref{fig:runtime}. In the class incremental learning setting, the model CCL-FP outperforms all compared methods on corresponding benchmarks by great margins and CCL-FP+ further obtains slight improvement except on 
S-Tiny-ImageNet. The regularization-based method SI fails in the class incremental setting on S-CIFAR-10, S-CIFAR-100 and S-Tiny-ImageNet and just gains negligible improvement on S-MNIST compared to rehearsal-based methods, suggesting that merely regularizing the parameter changes is not sufficient to overcome forgetting.  With a relatively small buffer of size 200, most of rehearsal-based methods are barely satisfactory especially on the complicated datasets such as S-CIFAR-100 and S-Tiny-ImageNet. The task incremental learning is generally considered as the easiest continual learning scenario, where all baselines performs fairly well on S-MNIST because of its 
simplicity. On three other datasets, some rehearsal-based methods underperform in the setup of the small buffer size and single epoch training, whereas our models still gain outstanding performance for all corresponding benchmarks. It is noted that there is no substantially difference between CCL-FP and CCL-FP+, suggesting that the supervised contrastive loss is removable in this setting. For domain incremental learning we have two benchmarks where CCL-FP+ is consistently better than CCL-FP, confirming the importance of supervised contrastive loss for this setting. 

In Figure \ref{fig:forgetting}, we show the forgetting measure on three selected datasets - S-MNIST in class incremental learning, S-CIFAR-10 in task incremental learning and R-MNIST in domain incremental learning. CCL-FP+ consistently produces the best performance compared to other CL competitors, demonstrating the outstanding ability of our model to overcome catastrophic forgetting even with a very small buffer. 

Figure \ref{fig:runtime} shows the run time (training + inference) on S-CIFAR-10, S-Tiny-ImageNet and R-MNIST datasets and we can observe that our model CCL-FP+ is consistently computational efficient over different benchmarks.

Additionally, we further study the impact of buffer size in terms of average accuracy by evaluating ER, HAL and CCL-FP+ on S-CIFAR-10, S-CIFAR-100 and R-MNIST datasets with a range of different buffer size. The results are reported in Table \ref{tab:buffer}, in which we can see that the average accuracy improves with the increase of the buffer size and our model consistently outperforms ER and HAL on three selected datasets for a wide range of buffer sizes.

\subsection{Ablation Study}
We conducted an ablation study to explore the impact of each component in our proposed model on S-MNIST and S-CIFAR-10 and R-MNIST datasets with default buffer size of 200. The results are reported in Table \ref{tab:ablation}. 
Our
models have been empirically demonstrated to be superior to the baseline ER with reservoir buffer on all benchmarks in Table \ref{tab:acc}, which confirms the effectiveness of proposed components in our models. 
Here we will investigate and analyze the specific effect of each component in different continual learning scenarios:
$w\neq 0$ indicates the inclusion of the feature propagation component, 
$\alpha\neq 0$ indicates the inclusion of the contrastive representation rehearsal component $\mathcal{L}_{cl}$,
and 
$\beta\neq 0$ indicates the inclusion of the supervised contrastive replay component $\mathcal{L}_{scl}$.

In the class incremental learning, we observe that the feature propagation ($w\neq 0$) yields remarkable performance gains compared to 
the other
two components, which is approximately $7\%$ on S-MNIST and $6\%$ on S-CIFAR-10. The other two components contribute in total about $5.5\%$ gains on S-MNIST and $1.3\%$ on S-CIFAR-10. In the task incremental learning, there is no prominent gains from a certain component. By considering the results in Table \ref{tab:acc} we conclude that the supervised contrastive loss ($\beta\neq 0$) achieves marginal performance gains so it is removable in this setting. In the domain incremental learning, the supervised contrastive loss appears to be particularly important compared to two other components, which produce about $1.3\%$ performance gains on R-MNIST. In addition, as shown in Table \ref{tab:acc}, embracing the supervised contrastive loss into the model can obtain about $2.3\%$ performance gains on the P-MNIST benchmark.

\section{Conclusion}
In this paper, we proposes an effective and also computational efficient continual learning method that works for all the CL scenarios. Instead of pure experience replay training, we first re-represent the example embeddings by incorporating the information of previous representation space via feature propagation and the model is then trained on the modified example embeddings. Moreover, to largely preserve the representation space from drastical changes when experiencing new tasks, we 
encourage the current example embeddings to approach the previous corresponding ones by a contrastive loss whereby the model is expected to keep a competent memory of what has learned in the past and overcome the problem of catastrophic forgetting after being exposed to new tasks. Furthermore, a supervised contrastive loss is leveraged in the model training to explicitly encourage the examples of the same class to cluster closely in representation space and meanwhile push example embeddings from different classes to be far apart.
The extensive experiments demonstrated the superiority of our models over a group of continual learning methods on six standard continual learning benchmarks 
in all three CL scenarios.

\bibliography{paper}

\end{document}